\documentclass[letterpaper, 10 pt, conference]{ieeeconf} 
\IEEEoverridecommandlockouts
\usepackage{graphicx}
\usepackage{amsmath,amssymb,mathtools}
\usepackage{float}

\usepackage{booktabs}     
\usepackage{tabularx}     
\usepackage{array}        
\newcolumntype{Y}{>{\raggedright\arraybackslash}X} 
\usepackage{multirow}
\usepackage{makecell}
\usepackage{adjustbox}
\usepackage{siunitx}
\usepackage{microtype}

\usepackage[caption=false,font=footnotesize]{subfig}

\usepackage[table]{xcolor}
\usepackage{pifont}
\usepackage{tabularx, array, booktabs, makecell}
\newcolumntype{C}{>{\centering\arraybackslash}X}

\usepackage{cite}

\graphicspath{{pic/}}

\providecommand{\citet}[1]{\cite{#1}}
\providecommand{\citep}[1]{\cite{#1}}

\usepackage[linesnumbered,ruled,vlined]{algorithm2e}
\SetKwInput{KwIn}{Input}
\SetKwInput{KwOut}{Output}
\SetAlCapNameFnt{\footnotesize}
\SetAlCapFnt{\footnotesize}
\usepackage[hidelinks]{hyperref}
\hypersetup{
  pdfauthor={},
  pdftitle={n}
}

\title{\LARGE \bf U-ARM: Ultra Low-cost General Teleoperation Interface for Robot Manipulation}
\author{Yanwen Zou,\; Zhaoye Zhou,\;Chenyang Shi,\;Zewei Ye,\;Junda Huang\textsuperscript{$\dagger$},\;
Yan Ding\textsuperscript{$\dagger$},\;
Bo Zhao \\ 
\textsuperscript{1}School of AI, SJTU,\; 
\textsuperscript{2}EvoMind Tech,\; 
\textsuperscript{3}IAAR-Shanghai,\; 
\textsuperscript{$\dagger$}Independent Researcher \\
\texttt{yanwenkkz@gmail.com, bo.zhao@sjtu.edu.cn}
}

\begin{document}
\maketitle
\thispagestyle{empty}
\pagestyle{empty}

\begin{abstract}
One of the key bottlenecks in realizing General Embodied AI is the shortage of sufficient high-quality training data, which is costly to collect.
We propose \textit{U-Arm}, a low-cost and rapidly adaptable
leader-follower teleoperation framework designed to
interface with most commercially available robotic arms.
Our system supports teleoperation through three
structurally distinct 3D-printed leader arms that share
consistent control logic, enabling seamless compatibility
across diverse robot configurations. Compared with prior
open-source leader--follower interfaces, we further
optimize mechanical design and servo selection, achieving a bill of materials (BOM) cost of \$50.5 for the 6-DoF leader arm and \$56.8 for the 7-DoF version. To enhance usability, we mitigate the common challenge of controlling redundant degrees of freedom through mechanical and control optimizations. Experiments show that U-Arm attains higher data collection efficiency and competitive task success rates compared with a low-cost game-controller baseline across multiple manipulation scenarios. We also open-sourced real-world manipulation data collected with U-Arm. The project website is \url{https://github.com/MINT-SJTU/LeRobot-Anything-U-Arm}.
\end{abstract}


\section{Introduction}
Collecting large-scale high-quality manipulation data for dual-arm robots has long been a bottleneck in policy learning. Especially, for vision-language-action models (VLAs) \cite{black2410pi0,kim2024openvla,bjorck2025gr00t,yang2025egovla,3d-vla,liu2025hybridvla}, the preliminary scaling law has been verified, and realizing the general Embodied AI requires the collection of massive amounts of real-world data. Compared to simulation or human data, real-world robot data is the most directly applicable for training robust policies \cite{bjorck2025gr00t}. Among the various data collection methods, currently human demonstration remains the primary approach for acquiring such real-world data.

Prior research has introduced a range of demonstration interfaces, which can be broadly categorized into end-effector trajectory recording devices and leader-follower teleoperation systems. End-effector trajectory recording devices such as DexCap \cite{wang2024dexcap}, UMI \cite{chi2024universal}, and OpenTelevision \cite{cheng2024open}, are often lightweight and easy to use. However, the collected data can suffer from issues such as kinematic singularities, exceeding the robot’s workspace, insufficient precision or need complicated post-processing. In contrast, leader-follower teleoperation systems such as ALOHA \cite{zhao2023learning} and GELLO \cite{wu2024gello} enable intuitive and physically constrained demonstrations through mechanically isomorphic leader arms. These systems help ensure that collected trajectories are physically feasible and executable by the robot.

Nevertheless, adapting leader-follower teleoperation systems to different commercial robotic arms often requires extensive engineering work and the cost barrier remains significant. ALOHA's dual-arm system exceeds \$50,000, while GELLO—though 3D-printable and more reproducible—still relies on relatively expensive Dynamixel motors, with BOM for a single arm over \$270.

\begin{figure*}[t]
  \centering
  \includegraphics[width=.95\textwidth]{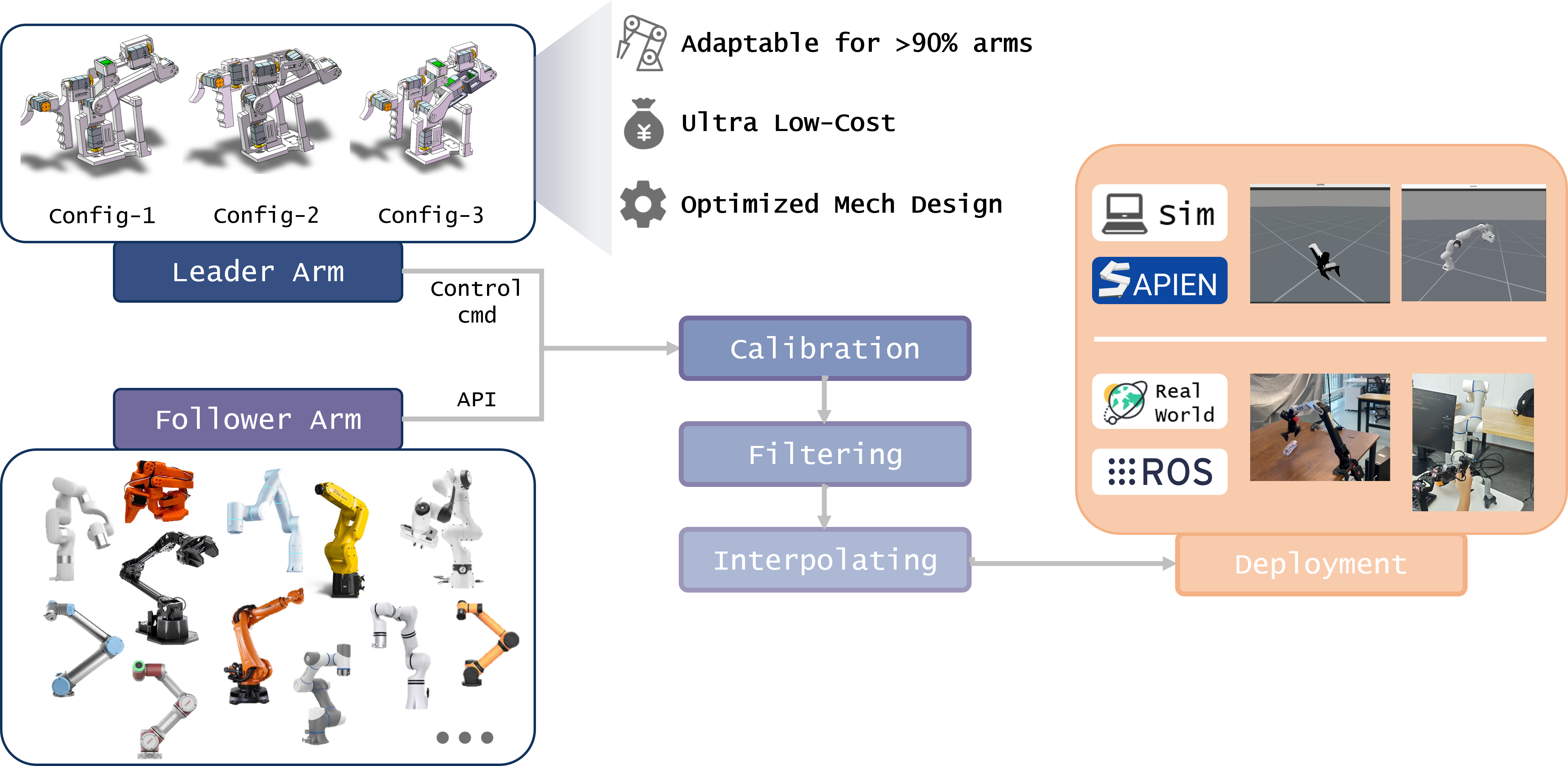}
  \caption{Overview of our proposed \textbf{U-Arm}, an ultra low-cost and general teleoperation interface for robot manipulation. 
    U-Arm provides three distinct mechanical configurations (Config-1, Config-2, Config-3), enabling adaptation to more than 95\% of commercial robotic arms. 
    The system supports both direct real-world teleoperation and simulation-based testing, with calibration, filtering, and interpolation modules ensuring stable and high-quality data collection. 
    Deployment is supported in SAPIEN as well as real-world robot platforms via ROS.}
  \label{fig:overview}
\end{figure*}

In this paper, we present an ultra-low-cost, user-friendly, and broadly compatible leader-follower teleoperation system that enables researchers to rapidly build customized data collection pipelines for their own robotic platforms. Fig.\ref{fig:overview} shows our system overview.  Specifically, we design three mechanically distinct leader arm configurations—two 6-DoF variants and one 7-DoF variant. Considering the structural constraints common to commercial robotic arms—namely, the need for satisfying Pieper's criterion in 6-DoF systems and anthropomorphic joint arrangements in 7-DoF systems—we observe that, although link lengths vary across platforms, the orientation and sequencing of joints generally follow standardized conventions. 

Following GELLO's idea that utilizing low-cost 3D-printed hardware to control commercial robotics arms, our leader arms are designed to provide sufficient mechanical isomorphism to ensure effective and compatible teleoperation across a wide range of commercial robots, while also incorporating a redesigned and optimized mechanical structure and components to enhance stability and further reduce overall cost to only $\sim\$50$
. Through a series of mechanical and control optimizations, we address the challenges introduced by redundant degrees of freedom in the leader arms. As a result, operators can achieve intuitive and effective control after only a few trials.

To summarize, we propose \textbf{U-Arm}, an ultra-low-cost leader--follower teleoperation system that is compatible with a wide range of commercial robotic arms. Our contributions are summarized as follows:

\begin{itemize}
    \item \textbf{System Design:} We develop and open-source an ultra-low-cost leader-follower teleoperation device that is compatible with most commercial robotic arms. All hardware designs are open-sourced.
    
    \item \textbf{Optimization Method:} We introduce a practical optimization strategy to address usability challenges caused by redundant degrees of freedom in high-DoF leader arms, improving control intuitiveness and efficiency.
    
    \item \textbf{Performance Gains:} Through data collection experiments, we demonstrate that U-Arm achieves \textbf{39\% higher demonstration efficiency} compared to existing JoyCon\cite{joycon_robotics}, another low-cost teleoperation devices, while maintaining a comparable success rate.
    
    \item \textbf{Community Support:} We provide simulation teleoperation examples based on SAPIEN\cite{xiang2020sapien}, as well as real-world manipulation datasets collected with U-Arm. 
\end{itemize}

\newpage
\section{Related Work}

\subsection{Manipulation Data Collection}

In the field of robotic manipulation, researchers are currently focused on the data-driven paradigm of imitation learning. As a result, various approaches to data collection have been explored to gather robot datasets. \emph{Learning from human videos} has recently emerged as a promising direction. Works such as GR00T\cite{bjorck2025gr00t} and EgoVLA\cite{yang2025egovla} leverage large-scale human-centric video datasets for policy training. At the same time, works like EgoMimic\cite{kareer2025egomimic}, Egozero\cite{liu2025egozero}, and Dexcap\cite{wang2024dexcap} propose more robot-oriented methods for collecting such human-centric video data. 

The advantage of this approach is that it enables knowledge transfer from the vast amount of human demonstration videos available online, and it holds the potential to scale up further with the development of more integrated hardware. However, this comes at the cost of introducing a significant human-robot embodiment gap. Consequently, such data can either be used primarily for pretraining large VLA models\cite{bjorck2025gr00t} or require heavy reliance on post-processing of videos and trajectory data to map human demonstrations into distributions more closely aligned with actual robot executions.

Another approach considered for scaling up robot data is \emph{ leveraging simulation platforms} such as Isaac Sim\cite{nvidia_isaac_sim}, MuJoCo\cite{conf/iros/TodorovET12}, and SAPIEN\cite{xiang2020sapien}. The most direct way to leverage simulation platforms for data augmentation is through scripted policies and domain randomization\cite{mandlekar2023mimicgen,jiang2025dexmimicgen}, which collect task demonstrations across different scenes and object placements. Some works have also utilized the capabilities of generative models to support the generation of new tasks and scenes, optimizing policies in the process\cite{wang2023robogen,li2024robot,hua2024gensim2}.

The greatest advantage of simulation environments is that researchers can easily access all ground-truth information within the scene for policy training and planning. Current rendering technologies have made it possible for simulation environments to achieve photo-realistic fidelity\cite{nvidia_isaac_sim}, closely resembling real-world scenarios. However, for high-dynamic, non-rigid, and contact-rich tasks with complex mechanical properties, simulators still face challenges in accurately simulating and executing policies, leaving a significant sim2real gap between simulated and real-world data.

\subsection{Teleoperation System for Bimanual Robot}

Embodiment and sim2real gap reflect the importance of collecting demonstrations directly from real-world through teleoperation. A key advantage of real-world teleoperation data lies in the separation between control and recording: teleoperation devices are typically used only for  control, while the data for learning is collected directly from the robot executing the task. This ensures that the recorded demonstrations share the closest distribution with deployment. However, real-world teleoperation inevitably involves human operators, making it the most costly and time-consuming method compared with simulation or vision-based data sources. Consequently, research on teleoperation devices mainly focused on reducing the cost of data collection, improving operator efficiency, and designing more user-friendly interfaces, all while ensuring high quality of the collected demonstrations.

 In recent years, a variety of teleoperation interfaces for bimanual robot manipulation have been developed to facilitate such data collection. Common devices include VR headsets~\cite{cheng2024open, xiong2025vision}, game controllers with joysticks, Spacemouse~\cite{liu2022robot}, multi-camera mocap systems~\cite{handa2020dexpilot, qin2023anyteleop, song2020grasping}, IMU-based controllers~\cite{wu2019teleoperation, laghi2018shared}, and leader-follower teleoperation systems~\cite{wu2024gello, zhao2023learning}. 

Most of these devices control the robot in end-effector space, which is then mapped to joint configurations through inverse kinematics (IK). In contrast, leader-follower teleoperation systems directly map the joint configurations of a leader arm to a follower robot, thereby avoiding singularity problems during motion and physically constraining the operator’s movements to ensure that collected demonstrations remain robot-executable.

Nevertheless, teleoperation systems must strike a balance between cost, usability, and control accuracy. In Table~\ref{tab:teleop_cost}, we summarize several representative teleoperation setups, comparing their hardware cost and qualitatively evaluating their usability for robot demonstration tasks.

\begin{table}[ht]
\centering
\caption{Cost and characteristics of representative teleoperation devices}
\begin{tabularx}{\linewidth}{l r X}  
\toprule
\textbf{Device} & \textbf{Price (USD)} & \textbf{Remarks} \\
\midrule
VR Headset (Meta Quest 3)~\cite{metaquest} & \$500 & May cause motion sickness \\
Space Mouse~\cite{spacemouse} & \$220 & Hard to operate bimanually \\
GELLO~\cite{wu2024gello} & \$270 & Uses Dynamixel motors, \$24 each \\
Game Controller (Joycon)~\cite{joycon_robotics} & \$20 & Difficult for dexterous manipulation \\
\textbf{Ours} & \textbf{\$50.5} & Sufficient for bimanual manipulation \\
\midrule
ALOHA~\cite{zhao2023learning} & \$24,000 & Identical leader-follower hardware design \\
DexPilot~\cite{handa2020dexpilot} & \$1700 & Faces occlusion problems \\
\bottomrule
\end{tabularx}
\label{tab:teleop_cost}
\end{table}

High-cost leader-follower teleoperation systems such as ALOHA benefit from advanced motor drivers and control algorithms, which enable gravity compensation. This significantly improves the overall usability of the system and alleviates the challenges associated with controlling redundant degrees of freedom. However, the prohibitive cost of such systems hinders their scalability and broader adoption in the research community. 

\subsection{Low-cost Teleoperation System}

GELLO~\cite{wu2024gello} can be considered as a low-cost alternative to ALOHA, aiming to support teleoperation across various commercial robot configurations. While game controller seems to be a sufficient cheap interface for simple manipulation tasks, we will demonstrate in the experiment section that it is less efficient than our device in actual use.

We got inspiration for developing an optimized low-cost general teleoperation interface from LeRobot project~\cite{cadene2024lerobot}, which emerged as a popular teleoperation interface for entry-level researchers. 
It introduces a low-cost 5-DoF desktop-scale robotic arm, primarily fabricated through 3D printing, with the goal of enabling users to quickly gain hands-on experience across the full pipeline of data collection, model training, and deployment. 
Despite its accessibility and popularity, the limited size and mechanical configuration of LeRobot hinder the generalization of trained models to the more commonly used 6-DoF and 7-DoF commercial robotic arms. 
Inspired by this observation, we argue that a universally adaptable, low-cost teleoperation interface that can seamlessly support diverse commercial robotic arm configurations is essential for scaling up high-quality robot demonstration datasets.

Based on GELLO, we identify two key points for further simplification and cost reduction: (1) For the leader arm, actuation at each joint is not strictly necessary. Since the leader arm only needs to be passively moved by the user while joint angles are recorded in real time, relatively expensive Dynamixel motors are not required. (2) Though mechanical isomorphism between the leader and follower arms is not essential, it is essential to ensure that the range of movement at each joint is appropriately constrained and maintain reasonable structural rigidity and durability for repeated use.

To this end, we propose U-Arm, a teleoperation interface that leverages ultra-low-cost motors for joint angle sensing, coupled with mechanical enhancements to address joint range constraints and the issue of redundant degrees of freedom. Through a series of validation experiments, we demonstrate that a leader-follower teleoperation architecture can be effectively implemented at a cost of only $\sim\$50$ per arm. Despite its low cost, the system offers sufficient mechanical compatibility and control fidelity to support teleoperation with the majority of commercial robotic arms available on the market.

\section{System Design}
\subsection{Hardware Design}



\begin{figure}[H]
  \centering
  \subfloat[Config 1 -- 6DoF]{%
    \includegraphics[width=.2\textwidth]{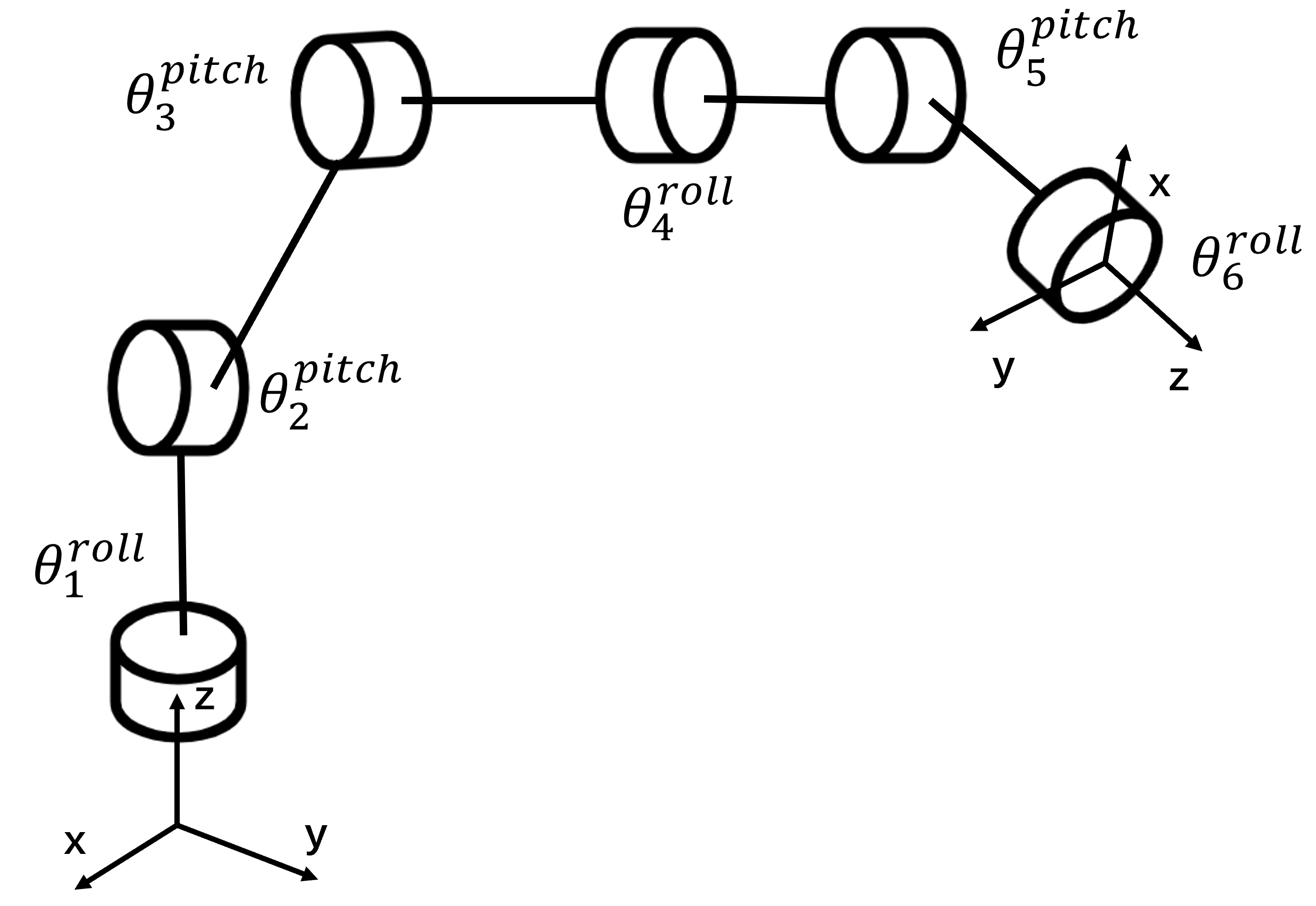}\label{fig:xarm}}
  \hfill
  \subfloat[Config 2 -- 6DoF]{%
    \includegraphics[width=.2\textwidth]{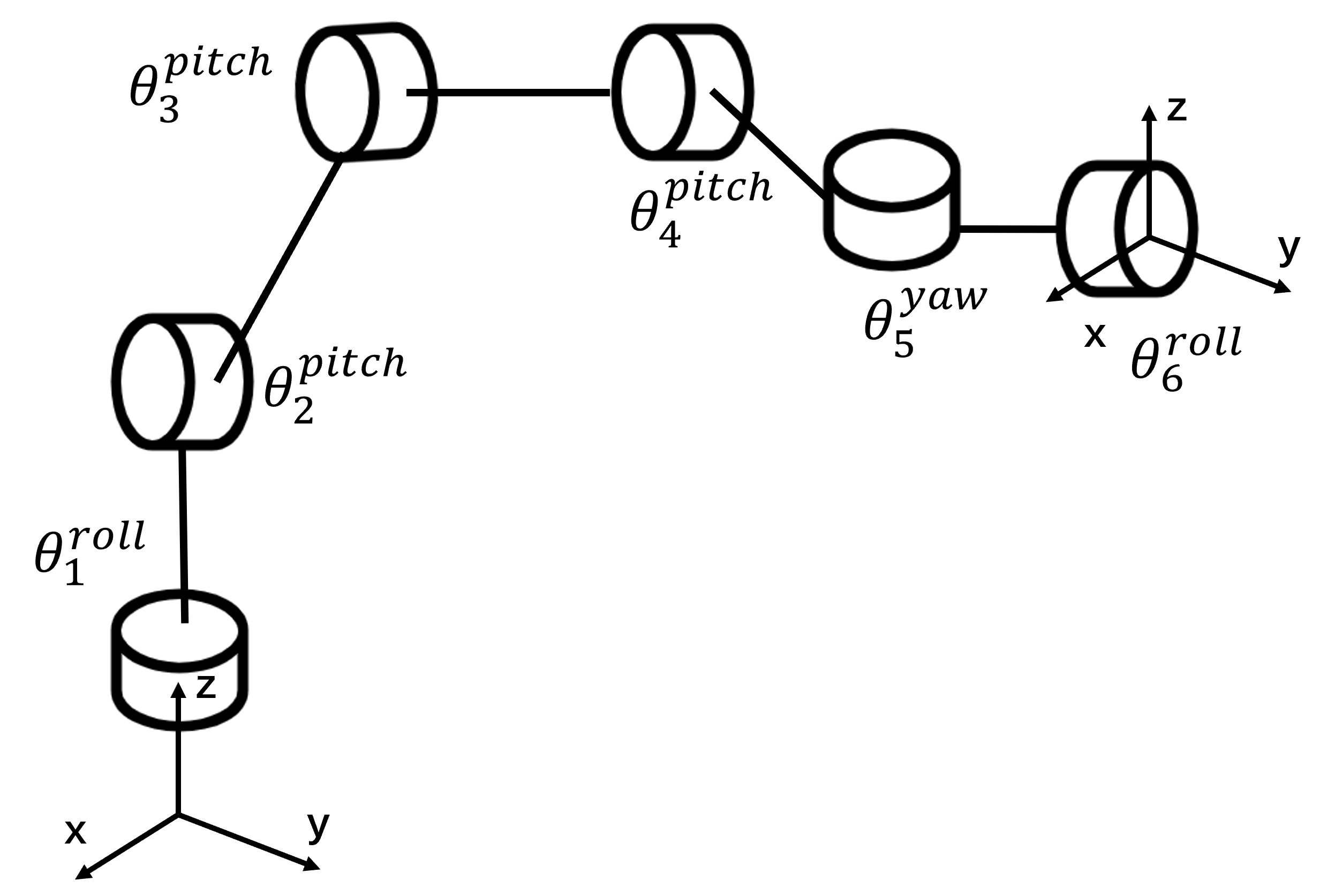}\label{fig:ur5}}
  \hfill
  \subfloat[Config 3 -- 7DoF]{%
    \includegraphics[width=.2\textwidth]{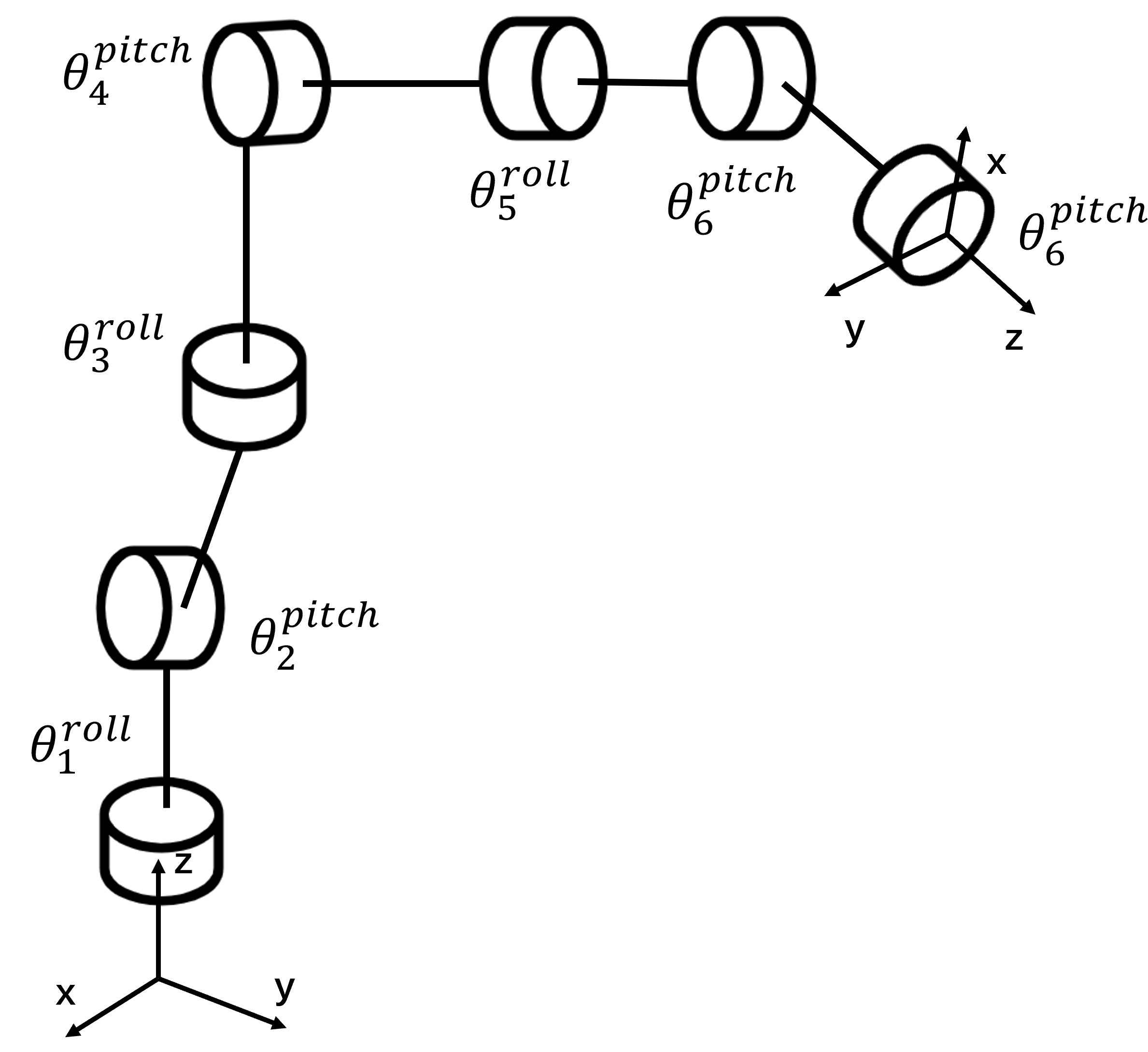}\label{fig:franka}}
  \caption{Joint axis arrangements of commercial robot arms.}
  \label{fig:pic1}
\end{figure}

\paragraph{Motivations} ~Most commercially available 6-DoF and 7-DoF robotic arms adopt one of only 3 standardized joint sequencing patterns, as illustrated in Fig.~\ref{fig:pic1}. We summarize in Table~\ref{tab:commercial_robots} a representative set of commercial robotic arms that are compatible with each of the three configurations. This design consistency arises from practical considerations such as satisfying Pieper's criterion, optimizing the dexterous workspace, and achieving anthropomorphic arm structures. While the exact link lengths of different arms may vary, their joint type and ordering typically follow one of these three common configurations.

For joint-space-controlled leader-follower teleoperation systems, we observe that it is not necessary for the leader arm to maintain a fixed link-length ratio relative to the follower arm. Instead, as long as the joint arrangement (i.e., the sequence of rotational axes) is the same, intuitive control can still be achieved. This is because, during teleoperation, the operator receives direct visual feedback from the motion of the follower arm. Therefore, the leader arm only needs to roughly convey the operator’s intended movement, rather than precisely replicate the follower’s kinematics.

\begin{table}[H]
  \centering
  \caption{Commercial robots compatible with each U-Arm configuration.}
  \label{tab:commercial_robots}
  \begin{tabularx}{\linewidth}{l X} 
    \toprule
    \textbf{U-Arm Config} & \textbf{Compatible Commercial Robots} \\
    \midrule
    Config-1 & xArm6, Fanuc LR Mate 200iD, Trossen ALOHA, Agile PIPER, Realman RM65B, KUKA LBR iiSY Cobot \\
    Config-2 & Dobot CR5, UR5, ARX RS5\textsuperscript{*}S, AUBO i5, JAKA Zu7 \\
    Config-3 & Franka FR3, Franka Emika Panda, Flexiv Rizon, Realman RM75B \\
    \bottomrule
  \end{tabularx}
\end{table}

\begin{figure}[H]
  \centering
  \subfloat[Config 1 -- 6DoF]{%
    \includegraphics[width=.2\textwidth]{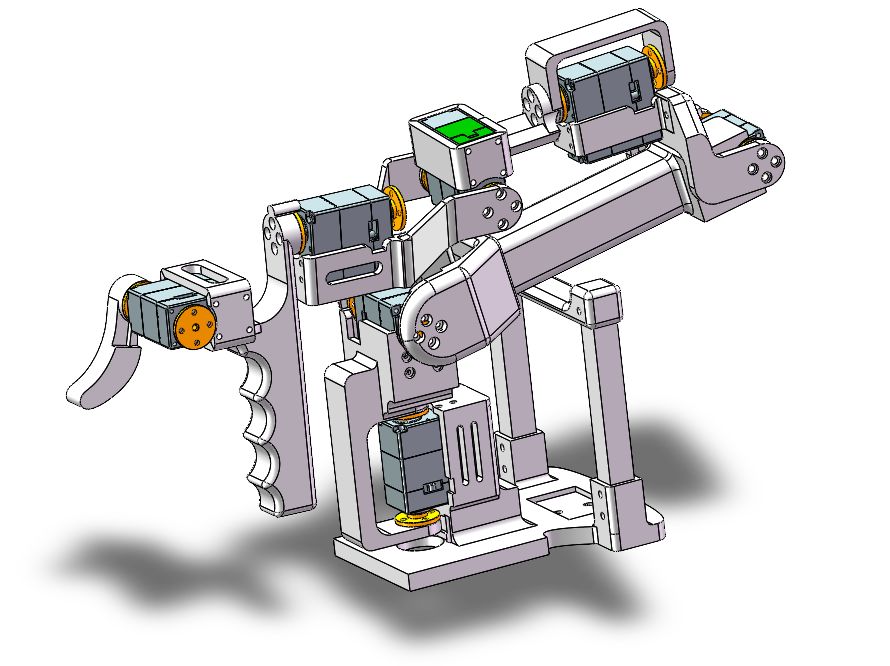}\label{fig:cad_cfg1}}
  \hfill
  \subfloat[Config 2 -- 6DoF]{%
    \includegraphics[width=.2\textwidth]{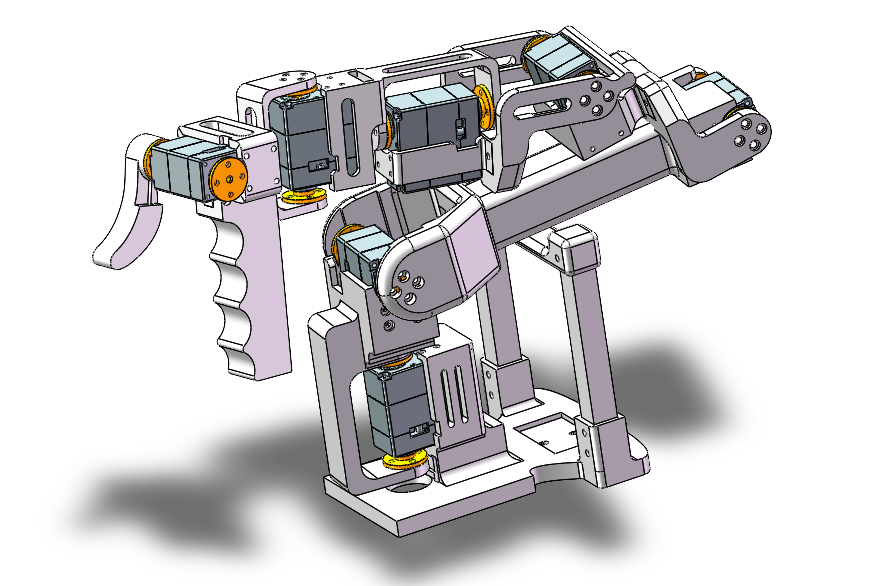}\label{fig:cad_cfg2}}
  \hfill
  \subfloat[Config 3 -- 7DoF]{%
    \includegraphics[width=.2\textwidth]{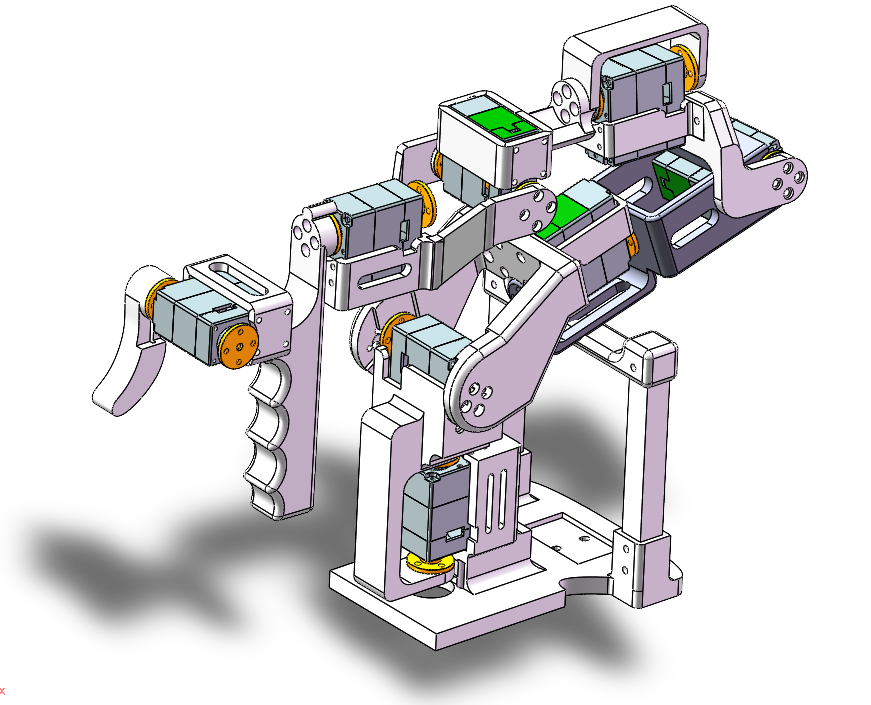}\label{fig:cad_cfg3}}
  \caption{CAD overviews of three configurations of U-Arm.}
  \label{fig:pic2}
\end{figure}

\paragraph{Mechanical Design} ~Corresponding to the configs presented above, the overview of mechanical design of the U-Arm is shown in Fig.~\ref{fig:pic2}. All parts are printed using PLA in our experiments. Considering PLA's strength, all components of the U-Arm have a minimum wall thickness of 4mm in order to ensure durability.

One common issue with low-cost 3D printed leader arms is the small surface area at the joint connection plates. Over time, these connections may loosen or even break due to the high radial reciprocating loads experienced by the first few joints. To address this, all joints in our design employ a dual-axis fixation method, which effectively mitigates this problem.

\paragraph{Motor Modification} ~In leader-follower teleoperation systems, joint resistance is a key factor that significantly affects the user experience. When the joint resistance of the leader arm is too high, operators find it difficult to perform smooth and continuous motions. On the other hand, if the resistance is too low, although the arm is easy to move intuitively, it may exhibit undesired passive motions in certain scenarios—particularly in Config-1 and Config-3.

For example, when the leader arm is extended horizontally near the boundary of its workspace, Joint 3 in Config-1 and Joint 4 in Config-3 tend to fall abruptly under gravity. While such instability can be addressed using gravity compensation algorithms with force-controlled motors, this would substantially increase system cost. GELLO partially mitigated this issue by adding rubber bands to selected joints to introduce passive resistance. 

In this work, the Zhongling servos we used were not originally designed for passive dragging or encoder-only applications. Their built-in gearboxes introduce excessive joint resistance, making smooth movement impractical. To address this, we adopt a more flexible mechanical solution. Since active actuation is not required in the leader arm, we disassembled the servos and removed their internal gears, retaining only the encoders for joint angle measurement. We then introduced damping by adjusting the tightness of the screws securing each joint axis, allowing us to finely control the resistance and stabilize the arm during teleoperation.

In addition, although the joint range varies across different commercial robotic arms, we intentionally designed the U-Arm—when used as the leader arm—with relatively narrower physical joint limits. These limits are sufficient to cover the requirements of typical tabletop manipulation tasks, while also enhancing the mechanical stability of the system by preventing extreme postures that could compromise structural integrity or cause unwanted joint behavior, as shown in Table ~\ref{tab:joint_range_config123}.

\begin{table}[!ht]
\centering
\caption{Joint range for U-Arm}
\renewcommand{\arraystretch}{1.2}
\begin{tabularx}{\linewidth}{c >{\centering\arraybackslash}X >{\centering\arraybackslash}X >{\centering\arraybackslash}X}
\toprule
\textbf{Joint} & \textbf{Config 1} & \textbf{Config 2} & \textbf{Config 3} \\
\midrule
Joint 1(base) & $[-87^\circ, 87^\circ]$ & $[-87^\circ, 87^\circ]$ & $[-87^\circ, 87^\circ]$ \\
Joint 2 & $[-75^\circ, 105^\circ]$ & $[-75^\circ, 105^\circ]$ & $[-70^\circ, 108^\circ]$ \\
Joint 3 & $[-180^\circ, 90^\circ]$ & $[-180^\circ, 90^\circ]$ & $[-70^\circ, 70^\circ]$ \\
Joint 4 & $[-72^\circ, 72^\circ]$ & $[-76^\circ, 50^\circ]$ & $[-180^\circ, 90^\circ]$ \\
Joint 5 & $[-122^\circ, 82^\circ]$ & $[-74^\circ, 74^\circ]$ & $[-72^\circ, 72^\circ]$ \\
Joint 6 & $[-115^\circ, 115^\circ]$ & $[-120^\circ, 120^\circ]$ & $[-122^\circ, 125^\circ]$ \\
Joint 7 & -- & -- & $[-115^\circ, 115^\circ]$ \\
\bottomrule
\end{tabularx}
\label{tab:joint_range_config123}
\end{table}
\footnotetext{For Config~2, we found that swapping joint~5 and joint~6 improves operator comfort. Because the corresponding robots use a cross-axis wrist structure for joints~4--6 with very short link lengths between them, the control logic can simply be inverted in the teleoperation software to preserve intuitive operation.}
\subsection{Algorithm Design}
\paragraph{Servo Encoder Adjustment and Calibration} ~This work uses joint angle mapping for teleoperation. The Zhongling servos used in our system have an encoder range of 0-270°, and exceeding this range can lead to unpredictable behavior. Since the gearboxes have been removed, the servos cannot correct their position through command feedback. Therefore, before installation, we manually adjusted the servos to a neutral position around 135°, ensuring that the leader arm does not exceed the encoder range during normal operation.

\paragraph{Calibration and Filtering} ~Even with identical joint configurations, different robotic arms exhibit significant structural differences, making it impossible to specify a unified initial position for all systems. Thus, during each teleoperation process, we initialize the leader arm near the follower arm's pre-defined initial pose. At the start of each demonstration, the follower robot firstly moves to its initial pose, then the leader arm module is initialized and taking control. During operation,the joint angles command sent by the leader-arm are filtered and interpolated to account for slight disturbances in the encoder readings, ensuring smooth and accurate motion despite mechanical variations between systems.

\begin{algorithm}[!t] 
\caption{Calibration and Filtering in Leader--Follower Teleoperation}
\KwIn{init leader angle $L_0$, init follower angle $F_0$, filtering threshold $\tau$, interpolation steps $N$}
\KwOut{control command sequence}

\For{each joint $j$}{
  $\Delta\theta \leftarrow L_j - L_{0,j}$\;
  $\theta^{\text{target}} \leftarrow \Delta\theta + F_{0,j}$\;

  \If{$\left|\Delta\theta\right| < \tau$}{\textbf{continue}\;}

  \For{$s \leftarrow 1$ \KwTo $N$}{
    $\theta^{\text{cmd}} \leftarrow \theta^{\text{cmd}} + \dfrac{\Delta\theta}{N}$\;
    send($\theta^{\text{cmd}}$)\;
  }
}
\end{algorithm}

\section{Experiments}

As a teleoperation device, the cost advantage of the U-Arm has already been discussed in the previous sections. 
In this section, we firstly adapt U-Arm to control different arms in SAPIEN~\cite{xiang2020sapien} to illustrate its applicability. Then we compare the data collection efficiency and reliability of the U-Arm with another representative low-cost teleoperation system, a standard game controller (Joycon), in real world.

Specifically, in real-world experiments, we use the U-Arm (Config-1) and the Joycon to teleoperate the Xarm6 robotic arm in a series of real-world tasks. 
During each experiment, we record the task success rate (emergency stop being counted as a failure) and the corresponding data collection time.

\subsection{Simulation Environment Adaption}
We adapt U-Arm to ManiSkill ~\cite{taomaniskill3}, which is built upon SAPIEN~\cite{xiang2020sapien}. This environment provides a safe setting for users to test teleoperation with U-Arm in different simulation scenes. Before operating the real robot, users can verify potential issues like errors in joint angle mapping and get easy access to teleoperation without the need of a real-world robot. Additionally, users can also collect several demonstrations in this setting for utilizing emerging data scaling method like MimicGen~\cite{mandlekar2023mimicgen}.

We set up the simulation environment in an indoor setting with various desktop items, where we test the capacity of the U-Arm to teleoperate robot arms to accomplish basic tabletop manipulation tasks and prove the cross-model generalization of U-Arm. 

\begin{figure}[ht]
    \centering
    \includegraphics[width=.95\columnwidth,trim=10 8 10 8,clip]{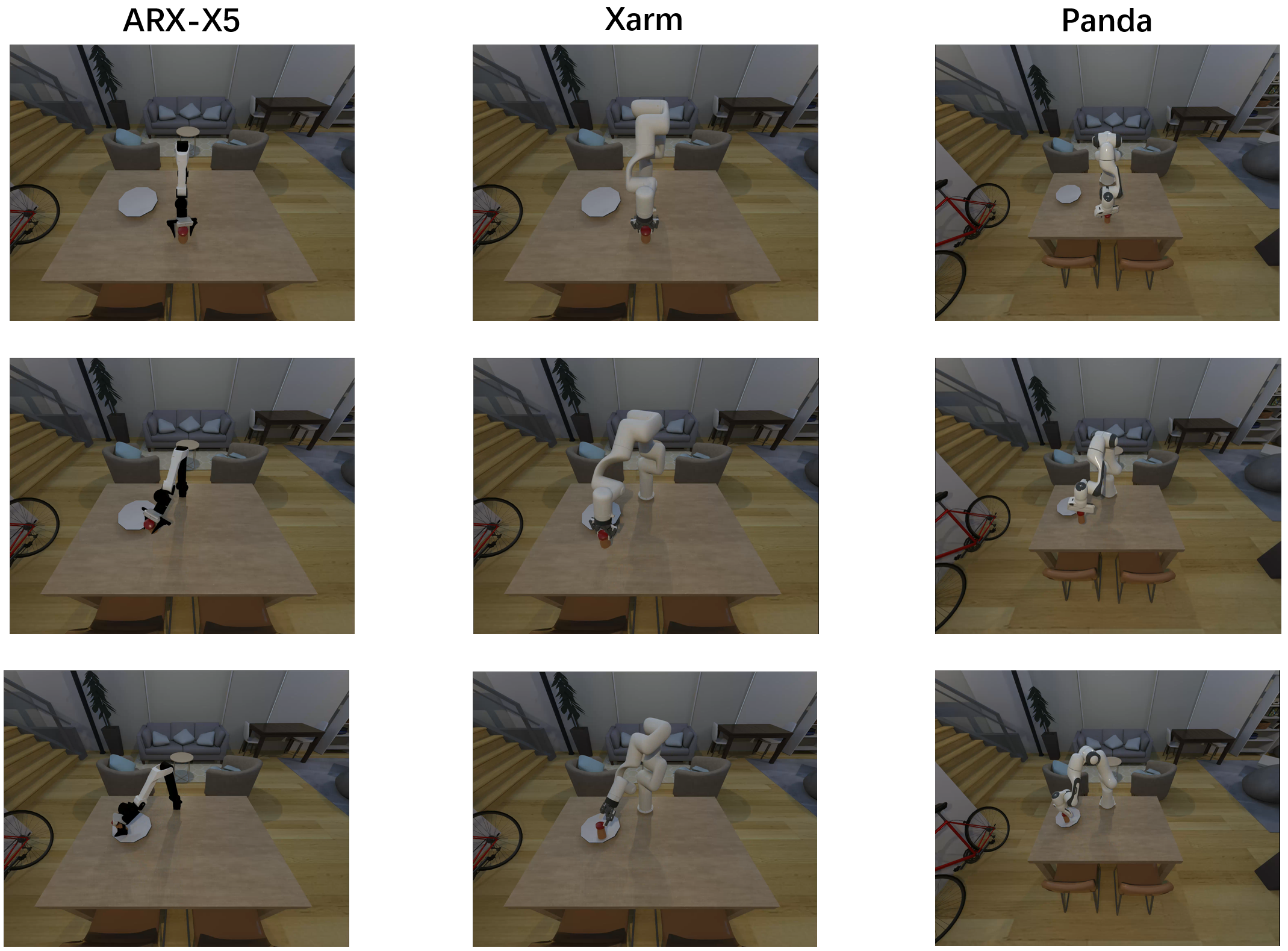}
    \caption{We setup three different robotic arms (ARX-X5, XArm, Panda) for teleoperation example in simulation, corresponding to three U-Arm configs.}
    \label{fig:sim_task}
\end{figure}

We provide simulation examples for controlling 7 different robot arms at our website, including ARX-X5, Xarm6, SO-100, Panda. Some examples are shown in Fig.~\ref{fig:sim_task}. 







\begin{figure*}[t]
    \vspace{3ex}
    \centering
    \includegraphics[width=.85\textwidth,trim=10 8 10 8,clip]{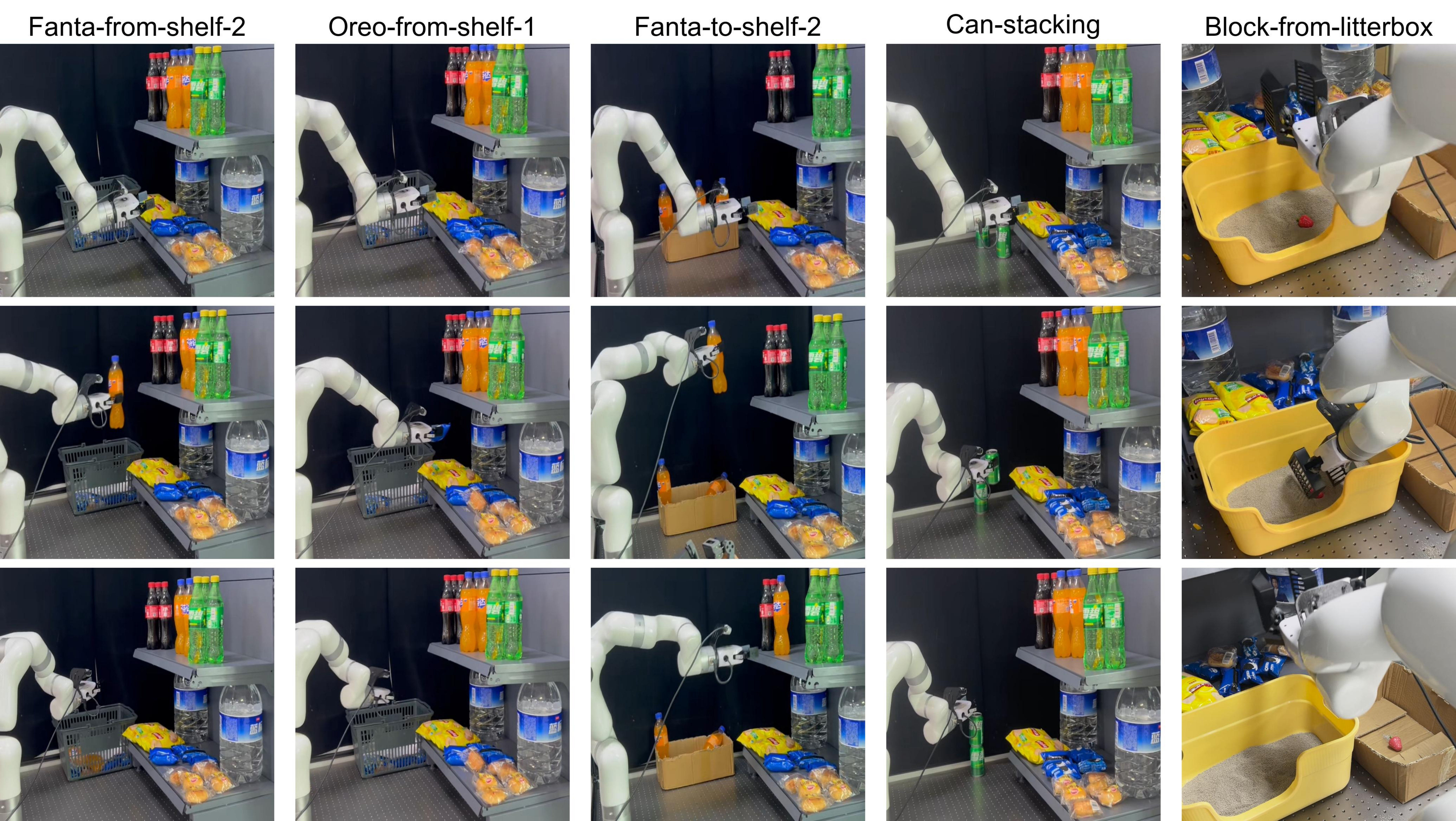}
    \caption{Real-world manipulation tasks performed with U-Arm Config2 and XArm6. These tasks include large-range movement  fine-grained operations and tasks that demand dexterity. Time cost and success rate are recorded for each demonstration.}
    \label{fig:real_task}
\end{figure*}

\subsection{Real-world Task Setup}
We design five representative real-world manipulation tasks Fig.~\ref{fig:real_task}:

\begin{enumerate}
    \item \textbf{Fanta-from-shelf-2:} Move a Fanta bottle from the second layer of the shelf to the basket.
    \item \textbf{Oreo-from-shelf-1:} Move an Oreo package from the first layer of the shelf to the basket.
    \item \textbf{Fanta-to-shelf-2:} Pick a Fanta bottle from a carton box and place it back onto the second layer of the shelf.
    \item \textbf{Can-stacking:} Stack a soda can on top of another can.
    \item \textbf{Block-from-litterbox:} Grasp a building block from the cat litter box.
\end{enumerate}

\subsection{Real-world Experiment}
Table~\ref{tab:realworld_results} summarizes the performance of the U-Arm and the Joycon across the five real-world tasks.
Overall, the U-Arm achieves consistently shorter data collection times across all tasks, while maintaining comparable success rates to the Joycon.

This advantage can be explained by the nature of tabletop tasks, which generally consist of two distinct phases:

\textbf{Large-range Movement}– the robot arm is quickly moved toward the vicinity of the target object.

\textbf{Fine-grained Adjustment} – once near the object, the operator slows down to carefully adjust the end-effector’s pose for precise manipulation.

While the Joycon, as an end-effector control device, is sufficient for performing fine adjustments, it is less natural for executing large, sweeping arm motions. In contrast, mechanically well-adjusted leader arm, due to its arm-mimicking control structure, enables operators to perform the first phase more intuitively and efficiently.
\begin{table}[ht]
\centering
\caption{Real-world teleoperation performance comparison}
\renewcommand{\arraystretch}{1.2}
\begin{tabular}{lcc@{\hskip 2em}cc} 
\toprule
\multirow{2}{*}{\textbf{Task}} & \multicolumn{2}{c}{\textbf{U-Arm (Config-1)}} & \multicolumn{2}{c}{\textbf{Joycon}} \\
 & Time (s) & Success (\%) & Time (s) & Success (\%) \\
\midrule
Fanta-from-shelf-2    & 14.43 & 88.8\% & 27.85 & 94.0\% \\
Oreo-from-shelf-1     & 11.28 & 88.5\% & 22.23 & 100.0\% \\
Fanta-to-shelf-2      & 19.88 & 72.2\% & 31.90 & 60.0\% \\
Can-stacking          & 20.93 & 39.6\% & 31.35 & 64.0\% \\
Block-from-litterbox  & 21.99 & 90.0\% & 31.89 & 96.0\% \\
\midrule
\textbf{Average}      & 17.70 & 75.8\% & 29.04 & 83.0\% \\
\bottomrule
\end{tabular}
\label{tab:realworld_results}
\end{table}

\noindent \paragraph{Observations}  
Across all tasks, the U-Arm demonstrated a 39\% reduction in operation time compared to the Joycon, without a substantial decrease in success rate. This improvement can be attributed to its more intuitive control scheme. 
In particular, for tasks involving richer variations in motion (e.g., grasping a Fanta and placing it on a shelf), the one-to-one mapping between the leader and follower arms in the U-Arm enabled smoother execution. 

It is important to note that the U-Arm achieved a lower success rate than the Joycon in tasks requiring fine precision, such as can stacking. This is mainly because when using the Joycon, operators can instantly halt the robot arm’s movement simply by releasing the joystick, thereby reducing unintended errors. In contrast, most teleoperation devices, including U-Arms, continuously transmit the operator’s hand motions to the robot side in real time, which inherently increases the likelihood of small mistakes during delicate manipulations. Nevertheless, we consider this reduction in success rate to be an acceptable trade-off when compared with the substantial gains in operation efficiency. 

\begin{figure}[ht]
    \centering
    \includegraphics[width=0.99\columnwidth,clip]{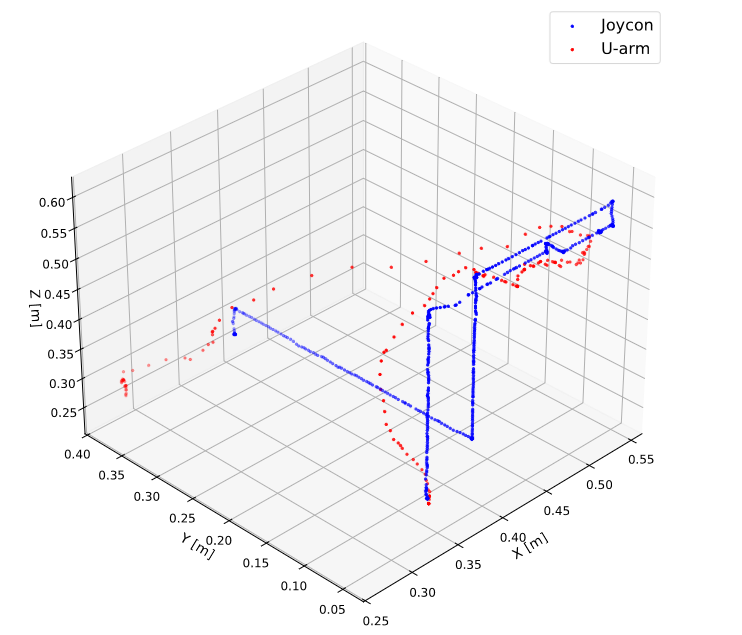}
    \caption{Comparison of end-effector trajectories collected using the Joycon (blue) and the U-Arm (red) during real-world manipulation tasks.}
    \label{fig:traj_comparison}
\end{figure}

\noindent
\paragraph{Trajectory Smoothness} 
As illustrated in Fig.~\ref{fig:traj_comparison}, the end-effector trajectories obtained from the U-Arm interface follow smoother and more continuous paths. 
In contrast, trajectories collected with the Joycon controller typically consist of piecewise, axis-aligned movements, reflecting the discrete nature of joystick inputs. 
This observation suggests that leader-follower teleoperation system induces a trajectory distribution that is closer to natural human manipulation. 
Such a property is particularly desirable if one aims to co-train policies with other data sources (e.g., ego-centric human demonstrations\cite{wang2024dexcap}), since the data distribution collected via the U-Arm better aligns with the statistics of human motion. 

\begin{figure}[ht]
    \centering
    \includegraphics[width=0.9\columnwidth,trim=5 5 0 5,clip]{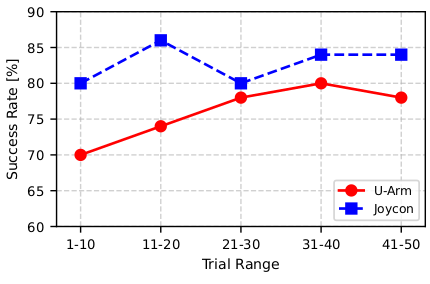}
    \caption{Success rate across different trial ranges for U-Arm (red) and Joycon (blue).}
    \label{fig:learning_curve}
\end{figure}

\noindent\paragraph{Analysis of Operator Proficiency}
It must be admitted that though some work are exploring shared autonomy in bimanual manipulation tasks\cite{luo2025human}, currently teleoperation system still highly rely on human labor and experience. To further investigate the effect of operator proficiency, we compare success rates across five trial ranges (1-10, 11-20, 21-30, 31-40, 41-50), as shown in Fig.~\ref{fig:learning_curve}. We observe that U-Arm shows a clear improvement after a short adaptation phase: while its initial success rate is slightly lower, it steadily increases and stabilizes around 80\% after 10-20 trials. 
In contrast, the Joycon exhibits relatively high success rates from the beginning, indicating that it is easier for operators to quickly adapt. 
Nevertheless, the U-Arm achieves comparable steady-state performance after a short learning period, demonstrating that its intuitive control scheme remains effective once operators become familiar with it.

\section{Conclusion}
In this work, we present U-Arm, an open-source, ultra-low-cost leader arm system designed for teleoperation. By providing three distinct mechanical configurations, U-Arm can be adapted to most commercial robotic arms on the market. To address the control challenges introduced by redundant degrees of freedom, we modified the  servos and applied an angle-mapping algorithm with filtering and calibration. These improvements enable for more stable and intuitive teleoperation. 

Experimental results demonstrate that U-Arm achieves 39\% higher data collection efficiency and comparable task success rates across multiple manipulation scenarios compared with Joycon, another low-cost teleoperation interface. Meanwhile, as a leader-follower teleoperation system, U-Arm performs trajectories that more closely reflect natural human motions compared with Joycon. This property makes U-Arm particularly advantageous when co-training policies with other human-centric data sources.

Despite these advantages, several known limitations remain. For instance, after extended use, servo connector looseness may occur due to the standard plug design; this could be mitigated in future versions by replacing them with soldered or more secure connectors. Additionally, the independent movement of Joint 5 in Config-1 and Config-3 depends on fine-tuning the tightness of mechanical fasteners, though our experiments suggest that these joints are seldom used in common tabletop manipulation tasks.

\section*{Acknowledgements}

We gratefully acknowledge Jiaqi Lu, Jie Yi, Yilei Zhong, Xinyu Chenyin for their assistance in hardware assembly, testing and experiment.

% ...

\bibliographystyle{IEEEtran}                 
\bibliography{iclr2025_conference}           

\end{document}